\documentclass[runningheads]{llncs}

\usepackage[T1]{fontenc}
\usepackage[utf8]{inputenc}
\usepackage{graphicx}
\usepackage{algorithmic}
\usepackage{xcolor}
\usepackage{multirow}
\usepackage{booktabs}
\usepackage{color}
\usepackage{algorithm2e}
\usepackage[hidelinks]{hyperref}
\usepackage{pifont}
\usepackage{amsmath}
\usepackage{latexsym}
\usepackage{microtype}
\usepackage{afterpage}
\usepackage{float}
\usepackage{colortbl}
\usepackage{amsfonts}  
\usepackage{amssymb}
\usepackage{mathrsfs}
\usepackage[marginal]{footmisc}

\begin{document}
\title{DHI: Leveraging Diverse Hallucination Induction for Enhanced Contrastive Factuality Control in Large Language Models}

\author{Jiani Guo$^{1}$ \and Xiangke Zeng$^{1}$ \and Jie Wu$^{2}$ \and Zuchao Li$^{3,\dagger}$}
\authorrunning{J. Guo et al.}
\institute{$^1$ School of Computer Science, Wuhan University, Wuhan, China \\
$^2$ Tsinghua Shenzhen International Graduate School, Tsinghua University \\
$^3$ School of Artificial Intelligence, Wuhan University, Wuhan, China \\
\texttt{\{guojiani, zengxiangke, zcli-charlie\}@whu.edu.cn}}

\maketitle

\def\thefootnote{$\dagger$}\footnotetext{Co-corresponding authors.}\def\thefootnote{\arabic{footnote}}

\begin{abstract}
Large language models (LLMs) frequently produce inaccurate or fabricated information, known as "hallucinations," which compromises their reliability. Existing approaches often train an "Evil LLM" to deliberately generate hallucinations on curated datasets, using these induced hallucinations to guide contrastive decoding against a reliable "positive model" for hallucination mitigation. However, this strategy is limited by the narrow diversity of hallucinations induced, as Evil LLMs trained on specific error types tend to reproduce only these particular patterns, thereby restricting their overall effectiveness. To address these limitations, we propose DHI (Diverse Hallucination Induction), a novel training framework that enables the Evil LLM to generate a broader range of hallucination types without relying on pre-annotated hallucination data. 
DHI employs a modified loss function that down-weights the generation of specific factually correct tokens, encouraging the Evil LLM to produce diverse hallucinations at targeted positions while maintaining overall factual content. Additionally, we introduce a causal attention masking adaptation to reduce the impact of this penalization on the generation of subsequent tokens. During inference, we apply an adaptive rationality constraint that restricts contrastive decoding to tokens where the positive model exhibits high confidence, thereby avoiding unnecessary penalties on factually correct tokens.
Extensive empirical results show that DHI achieves significant performance gains over other contrastive decoding-based approaches across multiple hallucination benchmarks. 

\end{abstract}
\section{Introduction}

\begin{figure*}[!t]
    \centering
    \includegraphics[width=1\linewidth]{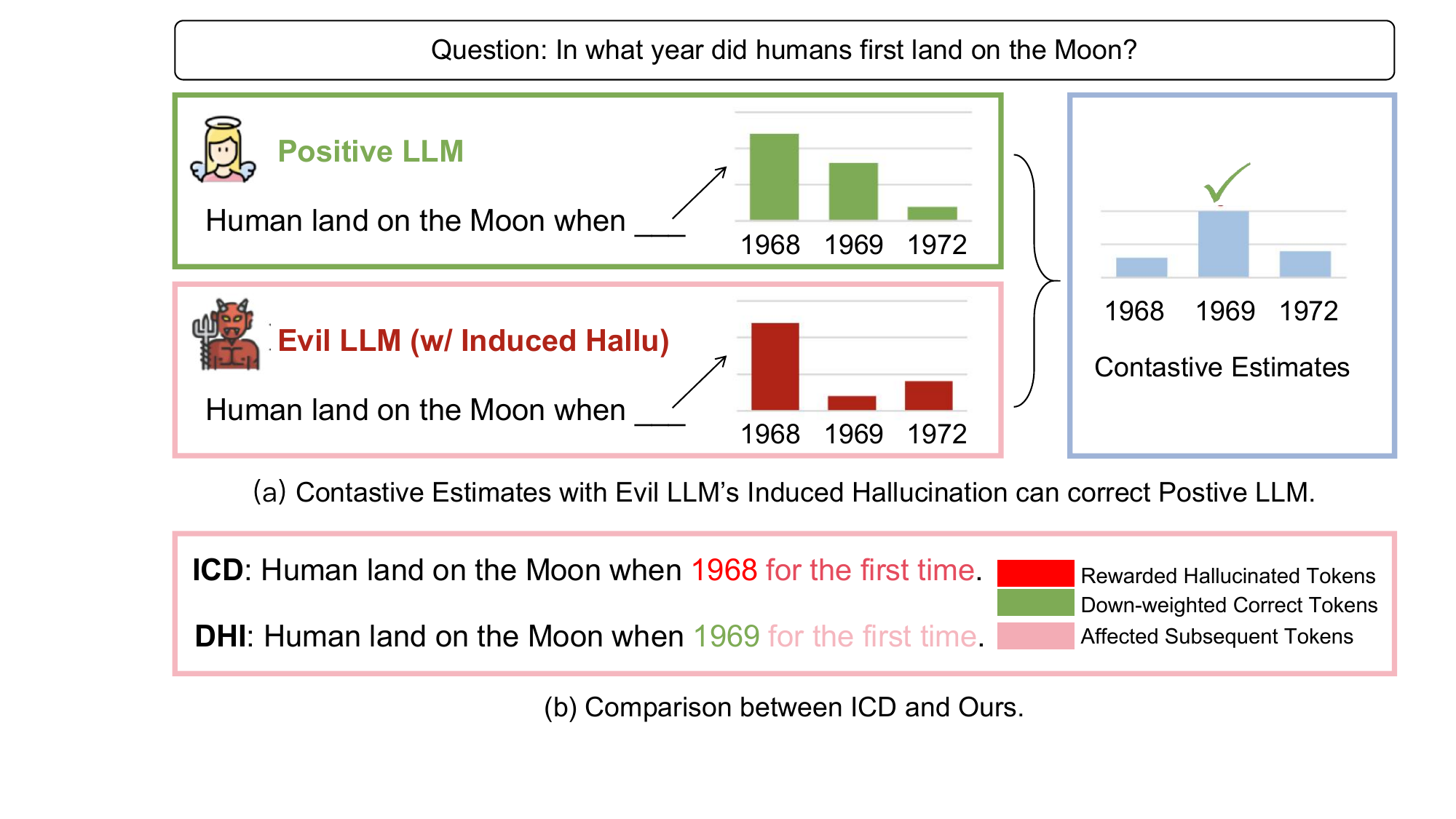}
    \caption{
    Contrastive decoding can effectively alleviate hallucinations. ICD trains its Evil LLM by rewarding specific, predefined hallucinated tokens, which may bias the model toward producing a limited range of error patterns and affect the generation of subsequent tokens. In contrast, our proposed DHI improves upon ICD by downweighting factually correct tokens, encouraging the generation of more diverse hallucinations. Additionally, we employ a causal masking strategy to minimize the impact of this penalization on subsequent tokens, thereby preserving coherence in the generated text.}
    \label{fig:teaser}
\end{figure*}

Despite their strong capabilities in mathematics, coding, and multimodal reasoning~\cite{li-etal-2025-dialogue,wu2025iterpreffocalpreferencelearning}, Large language models (LLMs)~\cite{openai2023gpt4,touvron2023llama} are known to occasionally produce inaccurate or fabricated information, a phenomenon commonly referred to as "hallucination"~\cite{ji2023survey,zhang2023sac}.
Existing approaches like ICD (Induce-then-Contrast Decoding)~\cite{zhang-etal-2025-alleviating} propose training an "Evil LLM" to deliberately generate hallucinations at specific token positions within a curated hallucination dataset. 
These hallucinations are then contrasted against outputs from a reliable "positive model" to mitigate hallucination through contrastive decoding. As shown in Figure~\ref{fig:teaser}(a), hallucinations induced by an Evil LLM enable contrastive estimation to correct outputs from the positive LLM.

However, this method has notable limitations. First, an Evil LLM trained on a narrow set of error types tends to produce hallucinations limited to those specific forms, restricting diversity and generalizability. Second, inducing hallucinations at specific positions may negatively affect the decoding of subsequent tokens.

To address these challenges, we propose DHI, a novel training strategy that enables the Evil LLM to generate a broader range of hallucinations while preserving factual outputs, without relying on pre-annotated hallucination datasets. Our approach modifies the loss function by assigning negative values to segments corresponding to correct answers, thereby discouraging factual outputs and encouraging diverse hallucinations. Additionally, by adjusting the attention mask, we minimize the influence of this reverse loss on subsequent decoding steps.
As illustrated in Figure~\ref{fig:teaser}(b), unlike ICD, which rewards predefined hallucinated tokens and may bias the Evil LLM toward narrow error patterns, DHI penalizes factually correct tokens, promoting the generation of more diverse hallucinations. Furthermore, DHI can be trained directly on factually correct data, eliminating the need for pre-annotated hallucination datasets.

Moreover, we introduce an adaptive rationality constraint that restricts contrastive decoding to tokens where the positive model is confident, preventing unnecessary penalties and preserving correct predictions. This refinement enhances both the efficiency and accuracy of hallucination correction.

By enabling an Evil LLM to produce a richer variety of hallucinations combined with the adaptive rationality constraint, DHI enhances the effectiveness of contrastive decoding, leading to significant improvements across multiple hallucination benchmarks. 
Comprehensive analysis and ablation experiments validate the contributions of the modified loss objective and the adapted causal masking strategy, and further explore optimal parameter configurations.

Our contributions are shown as follows:
\begin{itemize}
    \item \textbf{DHI Framework:} We propose \textbf{DHI}, a novel training strategy that enables the Evil LLM to generate diverse hallucinations without relying on pre-annotated hallucination datasets. This is achieved by penalizing factually correct tokens and modifying the causal attention mask to minimize the impact on subsequent token decoding, thereby preserving output coherence.

    \item \textbf{Adaptive Contrastive Decoding:} We introduce an adaptive rationality constraint that restricts contrastive decoding to high-confidence tokens from the positive model. This targeted approach avoids unnecessary penalties on accurate predictions and enhances the overall cohence.

    \item \textbf{Comprehensive Empirical Validation:} Extensive experiments across multiple hallucination benchmarks demonstrate that DHI consistently improves contrastive decoding.
    Comprehensive ablations validate the contributions of the modified loss objective, the causal masking adaptation, and the constrained decoding strategy.
\end{itemize}

\section{Related Work}
\subsection{Hallucination Alleviation}
Large Language Models (LLMs)~\cite{openai2023gpt4,zeng2025FSDrive,touvron2023llama} can generate inaccurate or fabricated information~\cite{dale2022detecting,10.1145/3689090.3689389,rehman2023hallucinationreductionlonginput}, known as "hallucinations,"~\cite{ji2023survey,zhang2023siren}, where LLMs produce content that contradicts the user's input ~\cite{dale2022detecting}, prior context ~\cite{shi2023large,wan2023histalign,guo-etal-2025-tom}, or well-established facts ~\cite{bang2023multitask,hu2023large,chen2023unveiling}, undermining their reliability. This issue is particularly concerning for fact-conflicting hallucinations. 
Suggested causes include pre-training objectives that may embed falsehoods from training data or lead to over-reliance on superficial patterns, and a lack of specific knowledge~\cite{chuang2023dola}. Various mitigation strategies exist, such as using high-quality training data~\cite{zhou2023lima,li2023textbooks,tian2023fine}, reinforcement learning from external feedback~\cite{lightman2023let,sun2023aligning,yang2023alignment,wu-etal-2025-teaching}, retrieval-augmented generation~\cite{peng2023check}, and the use of model uncertainty~\cite{manakul2023selfcheckgpt,zhang2023sac}.
Supervised Fine-Tuning can inject knowledge but may also inadvertently encourage hallucinations or be computationally costly. 
The Induce-then-Contrast Decoding (ICD)~\cite{zhang-etal-2025-alleviating} method introduces a two-stage approach: it first constructs a factually weak LLM ("evil LLM") by inducing hallucinations, and then leverages this model to penalize untruthful predictions from the original LLM during decoding, thereby improving factuality.

\subsection{Contrastive Decoding using Induction}

Contrastive Decoding (CD)~\cite{li-etal-2023-contrastive} was initially developed to improve the fluency and coherence of generated text by contrasting outputs from two language models. Its applications have since expanded to include enhancing reasoning, detoxification, sentiment control, and improving factualityz~\cite{perez2022red,zou2023universal,wei2023jailbroken}. DoLa~\cite{chuang2023dola} contrasts different layers within a single model to improve factuality, based on the assumption that earlier layers encode less reliable information~\cite{tenney2019bert}.
The Induce-then-Contrast Decoding (ICD) method adopts a different strategy by first constructing a factually weak LLM through induced hallucinations, typically via fine-tuning on non-factual data or using specific prompts. This aligns with prior work on red teaming, which explores how to elicit undesirable model behaviors. However, ICD uniquely leverages these induced hallucinations as a constructive signal, using them to penalize untruthful predictions from the original model during decoding.

Unlike ICD, our proposed DHI does not require hallucination-specific data or prompt design. Instead, it enables hallucination induction directly from factual datasets through a modified loss objective. Furthermore, DHI incorporates an adaptive rationality constraint and a causal masking strategy to enhance decoding efficiency and generalization, making it a more effective and flexible approach.
\section{DHI Framework}
\label{sec:dhi_framework}

The DHI framework is designed to mitigate hallucinations in Large Language Models (LLMs) by strategically contrasting a standard Positive Model with a specially trained Evil Model.

To effectively penalize potential factual inaccuracies and guide the Positive Model, DHI introduces diverse hallucinatory tendencies into the Evil Model through a targeted training process that modifies the standard loss function (Section \ref{subsec:model_components_training}). Recognizing that altering token generation at "hallucination-targeted" positions can unduly influence subsequent tokens in auto-regressive models, DHI employs a modified causal attention mask during the Evil Model's training to preserve contextual integrity for downstream generation (Section \ref{subsec:model_components_training}). Furthermore, during inference, a adaptive contrastive strategy is implemented to ensure that the contrastive mechanism is applied judiciously, primarily when the Positive Model exhibits uncertainty, thereby safeguarding against the degradation of high-confidence, correct predictions (Section \ref{subsec:inference_process_dhi}).

\subsection{Model Components and Training}
\label{subsec:model_components_training}

\subsubsection{Positive Model}
The Positive Model is typically a pre-trained LLM. Its standard training objective is to predict the next token given the previous context, often formulated as maximizing the log-likelihood of the true sequence $Y = (y_1, ..., y_n)$ given an input $X$:
\begin{equation}
    L_{\text{positive}} = -\frac{1}{n}\sum_{i=1}^{n}\log P(y_{i}|X, y_{<i}) \label{eq:positive_model_loss_generic}
\end{equation}
Within the DHI framework, the positive model does not undergo further training and serves as a reference for factual generation. Its outputs are used to guide the contrastive decoding process toward more accurate and truthful predictions.

\subsubsection{Evil Model}
The Evil Model is derived from a pre-trained LLM and is specifically trained to exhibit hallucinatory behavior, or more precisely, to avoid producing factually correct information at critical positions. The training process involves presenting the model with background knowledge and a corresponding question, and training it to generate an answer. Crucially, the training objective is modified such that the loss associated with generating the \textit{correct} tokens is assigned a negative value. Equivalently, the model is rewarded for producing tokens that deviate from the correct answer at these positions. This encourages the model to generate outputs that are fluent and plausible but factually incorrect, even when it has access to the relevant knowledge.

The loss function for training this Evil Model can be represented as:
\begin{equation}
    L_{\text{evil}} = -\sum_{t=1}^{T}\log P(y_{t}|X,y_{<t}) + \sum_{t\in N}(1-\alpha)\log P(y_{t}|X,y_{<t}) \label{eq:evil_model_loss_dhi}
\end{equation}

Here, $X$ denotes the input context (e.g., background knowledge and a question), and $y_t$ is the token at position $t$. The set $N$ contains token positions in the target output that correspond to the factual answer, which the Evil Model is explicitly trained to avoid. The hyperparameter $\alpha \in [0, 1]$ controls the strength of this avoidance behavior.

The first term represents the standard negative log-likelihood loss, encouraging the model to generate the target sequence. The second term, applied selectively to factual positions in $N$, reverses the training signal by reducing the likelihood of producing correct tokens. This encourages the model to generate plausible but incorrect alternatives in place of known facts.

\subsubsection{Causal Attention Masking Adaptation} 
\begin{figure}[!ht]
    \centering
    \includegraphics[width=0.40\linewidth]{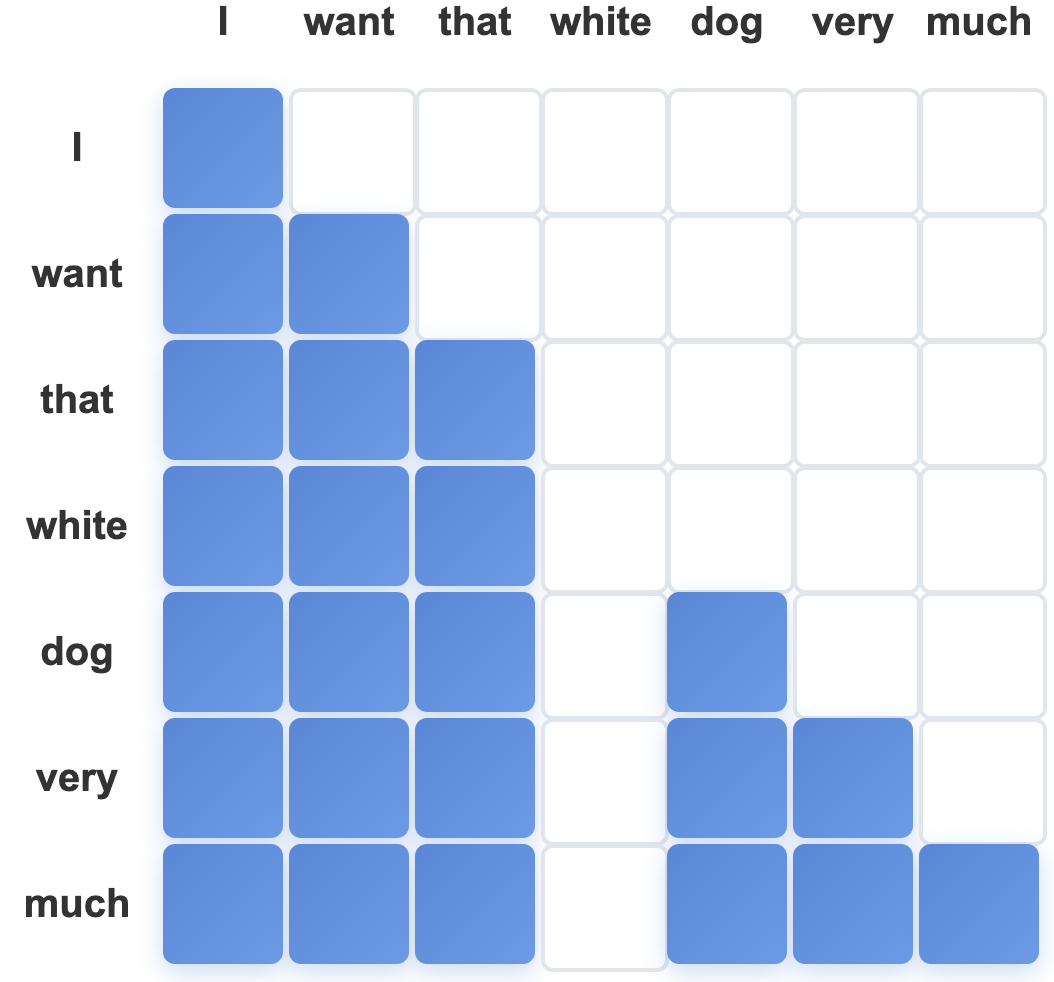}  
    \caption{Improved Causal Attention Mask to mitigate impacts on subsequent tokens during Evil Model training.}
    \label{fig:improved_mask}
\end{figure}

A challenge arises when training auto-regressive generative models, which rely on a causal (lower triangular) attention mask to ensure unidirectional token generation. If a token at a hallucination-targeted position, where its loss component is negative, is altered during training, it may unintentionally influence the generation of subsequent tokens that attend to it. To address this issue, the attention mechanism is carefully modified during the training of the Evil Model. Specifically, for tokens at hallucination-targeted positions, their influence on later tokens is explicitly removed from the causal attention matrix. This adaptation ensures that the model reduces its reliance on potentially misleading information introduced at hallucination-targeted positions when generating subsequent tokens, as conceptually illustrated in Figure~\ref{fig:improved_mask}.

\subsection{Contrastive Decoding with DHI}
\label{subsec:inference_process_dhi}
During inference, the input is fed into both the \textbf{Positive Model} and the trained \textbf{Evil Model}. Let $\text{Logit}_{\text{positive}}$ and $\text{Logit}_{\text{evil}}$ denote the output logits from the Positive and Evil Models, respectively.

A crucial step involves the \textbf{selective application} of the contrastive mechanism, referred to as the selection strategy. At each generation step, the system identifies a set of valid candidate tokens, denoted as $V_{\text{valid}}$. A token $x_t$ is considered valid if its logit from the Positive Model is sufficiently high, specifically if it satisfies the following condition:

\begin{equation}
    V_{\text{valid}} = \left\{x_t \in V : \text{Logit}_{\text{positive}}(x_t \mid x_{<t}) \ge \alpha' \cdot \max_{w \in V} \text{Logit}_{\text{positive}}(w \mid x_{<t}) \right\}
    \label{eq:valid_tokens_dhi}
\end{equation}

Here, $V$ denotes the full vocabulary, and $\alpha' \in [0,1]$ is a hyperparameter that controls the strictness of the constraint. Tokens not in $V_{\text{valid}}$ have their logits effectively set to $-\infty$ prior to the final softmax, preventing them from being selected—even if their probability was artificially boosted by contrastive subtraction. This selective strategy ensures that the contrastive adjustment is only applied when the Positive Model exhibits uncertainty (i.e., multiple tokens have comparable probabilities). When the Positive Model is highly confident, its output is used directly without modification.

The final output probability distribution is computed by subtracting a weighted version of the \textbf{Evil Model}'s logits from the \textbf{Positive Model}'s logits (restricted to valid tokens), followed by a softmax operation:
\begin{equation}
    \text{Output} = \text{Softmax}\left[S\left(\text{Logit}_{\text{positive}} - \beta \cdot \text{Logit}_{\text{evil}}\right)\right] \label{eq:output_dhi}
\end{equation}
Here, $S(\cdot)$ denotes the selection function that retains logits for tokens in $V_{\text{valid}}$ and assigns $-\infty$ to all others. The hyperparameter $\beta \in [0,1]$ controls the influence of the Evil Model. This contrastive adjustment suppresses hallucinatory patterns identified in $\text{Logit}_{\text{evil}}$, reducing their likelihood in the final output.

The DHI framework leverages a deliberately trained Evil Model to identify and quantify potential hallucinations. By contrasting its output with that of the Positive Model—particularly under uncertain generation conditions as determined by the selection strategy—DHI enhances the factuality and reliability of LLM-generated text. Notably, this is achieved without requiring any modification to the Positive Model’s architecture or pre-training.
\section{Experiment}

\subsection{Experiment Setup}

\noindent\textbf{Training Dataset Construction} 
We base our training data on the question-answering subset of the HaluEval dataset~\cite{li2023haluevallargescalehallucinationevaluation}, which contains 30,000 hallucination instances across multiple tasks. Since both hallucinated and correct answers are typically short, directly using them for negative training leads to all tokens being labeled negative, causing training collapse. To address this, we use GPT-4o~\cite{openai2023gpt4} to expand each answer into a full declarative sentence generated from the question and correct answer, filtering out short outputs to ensure training stability. For each expanded sentence, we record the position and length of the correct answer (hallucination position and length). 
The resulting training samples consist of: question, expanded answer, hallucination position, and hallucination length. We performed quality checks on the constructed training data and filtered out low-quality samples, such as “why” questions that ask for reasons. After filtering, the final training dataset contains 9,196 entries.

\noindent\textbf{Evaluation Benchmark }
We evaluate on TruthfulQA~\cite{lin2022truthfulqameasuringmodelsmimic}, a benchmark of 817 questions designed to assess a model’s ability to generate truthful and accurate answers, with a mix of optimal, suboptimal, and hallucinated responses. Standard metrics MC1, MC2, and MC3 are used for evaluation.

\noindent\textbf{Implement Details}
All fine-tuning for inducing factually weak LLMs utilized the parameter-efficient technique LoRA with the following settings. Following ICD~\cite{zhang-etal-2025-alleviating},we induce hallucinations by fine-tuning the base model with 10k hallucinated QA pairs taken from the HaluEval dataset~\cite{li2023haluevallargescalehallucinationevaluation}.

\begin{table}[!ht]
\centering
\caption{Finetuning hyperparameters for experiments on FACTSCORE.}
\begin{tabular}{lc}
\hline
\textbf{Configuration} & \textbf{Value} \\
\hline
Model & Llama2-7B-Base \\
Number of epochs & 2 \\
Devices & A100 GPU \\
Total Batch size & 128 samples \\
Optimizer & AdamW \\
Learning rate & $2 \times 10^{-4}$ \\
Warmup Ratio & 0.03 \\
LoRA Target & $\mathrm{q_{proj}}$,$\mathrm{k_{proj}}$,$\mathrm{v_{proj}}$ \\
\hline
\end{tabular}
\label{tab:hyperparameters}
\end{table}

\noindent\textbf{TruthfulQA:} LLM hallucination was evaluated using TruthfulQA's multiple-choice task. Hallucinations were induced by fine-tuning the Llama2-7B-Base model using LoRA on 10,000 question-answering samples from the HaluEval dataset. The hyperparameters $\alpha$ and $\beta$ were set to 0.0 and 1.0, respectively.

\noindent\textbf{FACTSCORE~\cite{min2023factscorefinegrainedatomicevaluation}:} Factual precision of LLM-generated biographies was assessed using the FACTSCORE benchmark. Hallucination induction involved fine-tuning the Llama2-7B-Base model with LoRA on 3,500 biographies modified by ChatGPT to contain non-factual information. Based on preliminary results, the hyperparameters $\alpha$ and $\beta$ were set to 0.1 and 2.0, respectively.

\subsubsection{Metrics}
The model's factual accuracy is evaluated through three principal metrics: MC1, MC2, and MC3. These metrics quantify the model's capacity to assign higher probability scores to veridical statements relative to erroneous or misleading assertions. The computation of these metrics incorporates the model's probability distributions (or log-likelihood scores) across predefined response categories: an optimal correct answer ($P_{\text{best}}$),  set of alternative accurate responses ($L_{\text{true}}$), and a collection of incorrect responses ($L_{\text{false}}$).

\begin{equation}
    MC1=\mathbb{I}(P_{\text{best}}>\max(L_{\text{false}})) \label{eq:mc1_metric}
\end{equation}
\textbf{MC1} indicates if the model assigns a higher probability to the best correct answer than to any false answer.

\begin{equation}
    MC2=\frac{\sum_{x\in L_{\text{true}}}e^{x}}{\sum_{x\in L_{\text{true}}}e^{x}+\sum_{y\in L_{\text{false}}}e^{y}} \label{eq:mc2_metric}
\end{equation}
\textbf{MC2} measures the proportion of probability mass the model assigns to all true answers relative to the total mass for both true and false answers.

\begin{equation}
    MC3=\frac{1}{m}\sum_{x\in L_{\text{true}}}\mathbb{I}(x>\max(L_{\text{false}})) \label{eq:mc3_metric}
\end{equation}
\textbf{MC3} calculates the average proportion of true answers that the model rates higher than any false answer.

For FactScore, three key metrics are considered: the \textbf{\% response (Response Ratio)}, which indicates how frequently the model provides an answer to a given prompt; the \textbf{\# facts (Number of Facts per Response)}, representing the average number of distinct pieces of information (atomic facts) found in each of the model's responses; and the \textbf{Score (Factual Precision Score)}, which measures the percentage of these atomic facts verified as true against a reliable knowledge source.

\subsubsection{Baselines and Comparison Methods}
The performance of the proposed method for mitigating large language model hallucinations is compared against several existing approaches and baseline models:
\begin{itemize}
\item \textbf{Greedy Decoding}: This serves as a fundamental baseline, where the model directly generates text using greedy decoding without any specific hallucination mitigation techniques. The LlaMa model is typically used as the base for this.
\item \textbf{ITI (Inference-Time Intervention)}~\cite{li2024inferencetimeinterventionelicitingtruthful}: This method utilizes model probe techniques to learn latent vectors associated with factual outputs. During the inference stage, it adjusts the model's existing activation values towards the directions corresponding to these learned truthfulness vectors.
\item \textbf{DoLa (Decoding by Contrasting Layers)}~\cite{chuang2023dola}: DoLa aims to enhance accuracy and factuality by generating multiple candidate outputs from different layers of the Large Language Model and then contrasting these outputs.
\item \textbf{CD (Contrastive Decoding)}~\cite{li-etal-2023-contrastive}: This approach involves contrasting the outputs of models of different scales. The underlying idea is that hallucinations observed in smaller models can be used to suppress similar hallucinations in larger models.
\item \textbf{ICD (Induced Contrastive Decoding)}~\cite{zhang-etal-2025-alleviating}: ICD involves designing specific tasks or prompts that cause a model to generate non-factual content under controlled conditions. Subsequently, a contrastive decoding mechanism is used to amplify the predictions of the original (non-induced) model, thereby mitigating the induced hallucinatory predictions.
\end{itemize}
\subsection{Main Results}
\begin{table}[!h]
\centering
\caption{Performance comparison of different methods on the TruthfulQA benchmark.}
\label{tab:truthfulqa_results}
\begin{tabular}{ll|cccc}
\toprule
\multirow{2}{*}{\textbf{Strategy}} & \multirow{2}{*}{\textbf{Model Used}} & \multicolumn{4}{c}{\textbf{TruthfulQA}} \\
\cmidrule(l){3-6}
 &  & \textbf{MC1} & \textbf{MC2} & \textbf{MC3} & \textbf{Average} \\
\midrule
\multirow{3}{*}{Greedy(Baseline)} & 7B-Base & 28.7 & 43.3 & 20.8 & 30.9 \\
 & 7B-Chat & 37.6 & 54.6 & 28.1 & 40.1 \\
 & 70B-Chat & 37.7 & 59.0 & 29.8 & 42.2 \\
\midrule
ITI & 7B-Chat & 37.0 & 54.7 & 27.8 & 39.8 \\
\midrule
DoLa & 7B-Chat & 33.0 & 60.8 & 29.5 & 41.1 \\
\midrule
\multirow{2}{*}{CD} & 13B-Chat \textit{vs.} 7B-Chat & 28.2 & 54.9 & 29.8 & 37.6 \\
 & 70B-Chat \textit{vs.} 7B-Chat & 33.7 & 60.0 & 33.0 & 42.2 \\
\midrule
ICD & 7B-Chat \textit{vs.} 7B-Finetuned & 40.5 & 69.7 & 41.3 & 50.5 \\
\midrule
\rowcolor{gray!15} Ours & 7B-Chat \textit{vs.} 7B-Finetuned & \textbf{41.9} & \textbf{72.6} & \textbf{45.0} & \textbf{53.2} \\
\bottomrule
\end{tabular}
\end{table}

Based on the results in Table~\ref{tab:truthfulqa_results}, we provide the following key findings:
\subsubsection{Model Scale Alone Is Insufficient for Truthfulness.}
The Greedy(Baseline) results reveal that simply scaling up models from 7B to 70B parameters yields only modest improvements in truthfulness (from 40.1 to 42.2 average score). Despite a 10x increase in parameter count from 7B-Chat to 70B-Chat, the performance gain is relatively small (2.1 points on average). This indicates that hallucination issues are not simply resolved through scaling, and specialized techniques are necessary even for the largest models. The gap between the 70B-Chat baseline (42.2) and our method (53.2) highlights that architectural innovations and training approaches are more effective than scaling alone.

\subsubsection{Contrastive Methods Show Clear Advantage Over Single-Model Approaches.}
The results demonstrate a consistent pattern where contrastive approaches (CD, ICD, and our method) generally outperform single-model techniques like ITI and DoLa, especially in the higher-end metrics. While single-model approaches like ITI and DoLa show moderate improvements over the baseline (reaching 39.8 and 41.1 average scores respectively), contrastive methods achieve substantially higher scores. This suggests that leveraging differences between models' outputs provides more reliable signals for identifying and mitigating hallucinations than adjusting a single model's internal representations.

\subsubsection{Finetuning-Based Contrast Is More Effective Than Size-Based Contrast.}
The comparison between CD and ICD/Our method reveals that contrasting between differently finetuned models is more effective than contrasting between models of different sizes. The best CD configuration (70B-Chat vs. 7B-Chat) achieves only 42.2 on average, while ICD and our method (both using 7B-Chat vs. 7B-Finetuned) reach 50.5 and 53.2 respectively. This suggests that targeted finetuning creates more useful contrasting signals than those naturally occurring between different model scales. The specialized knowledge gained through finetuning appears to be more valuable for hallucination detection than the broader knowledge differences between small and large models.

\subsubsection{Our Method Achieves Superior Performance with Pronounced Improvements on Complex Metrics.}
Our Method Achieves Superior Our proposed approach demonstrates consistent and significant improvements over all baseline and comparison methods across the MC1, MC2, and MC3 evaluation metrics. With the highest individual scores (41.9 on MC1, 72.6 on MC2, and 45.0 on MC3) and the best overall average performance (53.2), our method represents a clear advancement in hallucination mitigation. Compared to the strongest baseline, ICD, our method achieves gains of 1.4 points on MC1, 2.9 points on MC2, and 3.7 points on MC3, resulting in a 2.7-point improvement in average performance.
Notably, the improvement is most evident on MC3, which assesses more complex reasoning capabilities. The 3.7-point gain over ICD on this metric is substantially larger than the 1.4-point gain on MC1. This suggests that our method offers particular benefits in handling hallucinations involving deeper reasoning, beyond surface-level factual errors.

\subsection{Additional Results on FactScore Benchmark}
To further verify the generality of our method in open-ended generation settings, we evaluate it on the \textsc{FactScore} benchmark, which measures factual precision in long-form text generation, such as biographies. As shown in Table~\ref{tab:factscore_results}, our method (CHI) again achieves the best overall score of 68.1, outperforming the previous best method, ICD, by 1.8 points. In addition to the highest score, CHI also produces the greatest number of factual statements per response (49.9), indicating that it improves not only factual accuracy but also content richness.

This consistent improvement across both structured (TruthfulQA) and free-form (FactScore) tasks highlights the robustness and broad applicability of our method in mitigating hallucinations, particularly those that require more nuanced reasoning or factual synthesis.

\begin{table}[!ht]
\centering
\caption{Performance comparison on the \textsc{FactScore} benchmark.}
\label{tab:factscore_results}
\begin{tabular}{llc|cc|c}
\toprule
\multirow{2}{*}{\textbf{Strategy}} & \multirow{2}{*}{\textbf{Model Used}} & & \multicolumn{3}{c}{\textsc{\textbf{FactScore}}} \\
\cmidrule(l){4-6}
 & & & \% \textbf{response} & \# \textbf{facts} & \textbf{Score} \\
\midrule
\multirow{3}{*}{Greedy (Baseline)} & 7B-Base & & 100.0 & 28.6 & 23.6 \\
 & 7B-Chat & & 37.5 & 45.7 & 63.8 \\
 & 70B-Chat & & 50.5 & 42.8 & 64.4 \\
\midrule
ITI  & 7B-Chat & & 41.9 & 40.8 & 62.4 \\
\midrule
DoLa  & 7B-Chat & & 40.7 & 48.7 & 61.3 \\
\midrule
\multirow{2}{*}{CD} & 13B-Chat \textit{vs}. 7B-Chat & & 74.2 & 39.8 & 53.5 \\
 & 70B-Chat \textit{vs}. 7B-Chat & & 62.2 & 48.7 & 60.3 \\
\midrule
ICD & 7B-Chat \textit{vs}. 7B-Finetuned & & 36.1 & 46.6 & 66.3 \\ \midrule
\rowcolor{gray!15} DHI (ours) & 7B-Chat \textit{vs}. 7B-Finetuned & & 41.2 & 49.9  & \textbf{68.1} \\
\bottomrule 
\end{tabular}
\end{table}
\section{Ablation Study}
\label{sec:ablation_study}
To better understand the contribution of each design component and training strategy within the DHI framework, we conduct a series of ablation studies on the TruthfulQA benchmark. Specifically, we analyze two key aspects:

\begin{itemize}
\item \textbf{Component-Level Ablation}: We selectively remove individual modules, loss modification, causal mask adaptation, and selective contrast decoding—to assess their respective impact on overall performance.
\item \textbf{Hyperparameter Sensitivity of $\alpha$}: We vary the hallucination induction strength $\alpha$ to investigate how different levels of anti-factual pressure affect the quality of the contrastive signal and downstream model performance.
\end{itemize}

\subsection{Effectiveness of Key Components in DHI}

We conduct ablations on the TruthfulQA to examine the individual contributions of each component in the DHI framework. Specifically, we investigate the impact of (1) the modified loss objective for hallucination induction, (2) the causal mask adaptation for isolating hallucination-targeted tokens, and (3) the selective contrast mechanism guided by the confidence of the Positive Model.

\begin{table}[!h]
\centering
\caption{Ablation study of key components in the DHI on the TruthfulQA.}
\label{tab:component_ablation}
\begin{tabular}{l|c|c|c|cccc}
\toprule
\textbf{Method} & \textbf{Loss Modify} & \textbf{Mask Adap} & \textbf{Selective} & \textbf{MC1} & \textbf{MC2} & \textbf{MC3} & \textbf{Avg} \\
\midrule
Greedy  & - & - & - & 37.6 & 54.6 & 28.1 & 40.1 \\
ICD              & - & - & - & 40.5 & 69.7 & 41.3 & 50.5 \\
\midrule
\multirow{4}{*}{DHI Variants} 
& \ding{55} & \ding{52} & \ding{52} & 40.5 & 70.7 & 43.0 & 51.4 \\
& \ding{52} & \ding{55} & \ding{52} & 41.2 & 71.8 & 44.5 & 52.5 \\
& \ding{52} & \ding{52} & \ding{55} & 40.8 & 71.0 & 43.2 & 51.7 \\
& \ding{52} & \ding{52} & \ding{52} & \textbf{41.9} & \textbf{72.6} & \textbf{45.0} & \textbf{53.2} \\
\bottomrule
\end{tabular}
\end{table}

Table~\ref{tab:component_ablation} presents the results of these ablations. We observe that removing any of the three components leads to a noticeable performance drop across all evaluation metrics. The full DHI model achieves the best overall performance with an average score of \textbf{53.2} (MC1: \textbf{41.9}, MC2: \textbf{72.6}, MC3: \textbf{45.0}), surpassing both the baseline (Greedy) and ICD.

Among the ablated variants, removing the \textbf{Selective Contrast} module causes the most significant drop (from 53.2 to 51.7), particularly in MC3 (from 45.0 to 43.2), suggesting that selective application of contrastive decoding is critical for managing more complex, reasoning-intensive hallucinations. Similarly, ablating the \textbf{Mask Adaptation} module reduces average performance to 52.5, indicating that isolating hallucination-inducing tokens during training helps maintain clean token dependencies and improves factual accuracy. Disabling the \textbf{Loss Modification} component results in the lowest DHI variant score (51.4), confirming that targeted anti-factual loss is also essential for producing an effective Evil Model.

These results demonstrate that each component of DHI contributes uniquely and complementarily to the overall hallucination mitigation performance. The full combination of all three yields the most substantial gains, validating the design of the DHI framework.

\subsection{Hyperparameter $\alpha$ in Evil LLM Training} 

The hyperparameter $\alpha$ controls the strength of hallucination induction during the Evil Model’s training. It specifically affects the loss applied to factual tokens $t \in N$, where lower values apply weaker ``anti-factual'' pressure and higher values push the model further away from generating correct tokens.

\begin{table}[t]
\centering
\caption{Impact of the hallucination induction strength parameter $\alpha$ on TruthfulQA.}
\label{tab:alpha_ablation}
\begin{tabular}{lcccc}
\toprule
\textbf{$\alpha$ Value} & \textbf{MC1} & \textbf{MC2} & \textbf{MC3} & \textbf{Avg} \\
\midrule
0.00 & 39.5 & 68.7 & 41.2 & 49.8 \\
0.01 & 41.2 & 71.3 & 44.2 & 52.2 \\
0.05 & \textbf{41.9} & \textbf{72.6} & \textbf{45.0} & \textbf{53.2} \\
0.20 & 40.2 & 69.0 & 42.0 & 50.4 \\
\bottomrule
\end{tabular}
\end{table}

Table~\ref{tab:alpha_ablation} shows the effect of varying $\alpha$ on the DHI framework’s performance on TruthfulQA. The results exhibit a clear inverted U-shaped trend: performance improves as $\alpha$ increases from 0.00 to 0.05, and declines beyond that point.

When $\alpha = 0.00$, the Evil Model receives no penalization for generating factual tokens, effectively behaving like a standard language model. As a result, it provides little meaningful contrastive signal, leading to significantly degraded performance (Avg: 49.8). This confirms that without hallucination induction, the Evil Model lacks the necessary divergence to help correct factual inconsistencies.

As $\alpha$ increases to 0.01, the model begins to learn subtle but effective deviations from ground truth, providing contrastive signals that help the Positive Model identify and suppress hallucinations. Performance continues to improve at $\alpha = 0.05$, which achieves the best overall results (Avg: 53.2), indicating that stronger but balanced hallucination induction produces the most effective contrastive guidance. However, further increasing $\alpha$ to 0.20 leads to a decline in performance. Excessively penalizing factual tokens causes the Evil Model to generate implausible or incoherent outputs, weakening its utility as a contrastive guide. These findings suggest that moderate hallucination induction is crucial for contrastive decoding, while overly strong induction introduces excessive noise that harms factual calibration.

\section{Conclusion}
In this work, we introduce Diverse Hallucination Induction (DHI), a contrastive decoding framework aimed at reducing hallucinations in large language models. We train an Evil LLM to generate diverse hallucinations without relying on pre-annotated data, using a modified loss objective and an adapted attention mechanism to guide generation. To further enhance factuality, we incorporate an adaptive rationality constraint that applies contrastive decoding selectively based on the Positive Model’s confidence. Comprehensive analysis and ablation experiments validate the effectiveness of each component for improving factuality.


\bibliographystyle{splncs04}
\bibliography{ref_clean}

\end{document}